\documentclass[letterpaper, 10 pt, conference]{ieeeconf}  

\IEEEoverridecommandlockouts                              

\usepackage{amsfonts,amssymb,amsmath}
\usepackage{algorithmic}
\usepackage{algorithm}
\usepackage{array}
\usepackage{caption}
\usepackage[caption=false,font=normalsize,labelfont=sf,textfont=sf]{subfig}
\usepackage{textcomp}
\usepackage{stfloats}
\usepackage{url}
\usepackage{verbatim}
\usepackage{graphicx}
\usepackage{cite}
\usepackage{siunitx}
\usepackage{makecell}
\usepackage{multirow}
\usepackage{booktabs}
\usepackage{amsmath}
\usepackage{amsfonts}

\graphicspath{{figures}}
\usepackage{xcolor}
\usepackage{diagbox}   
\usepackage{array}
\usepackage{siunitx} 
\usepackage[hidelinks]{hyperref}
\usepackage{tikz}
\usetikzlibrary{arrows.meta,positioning,fit,backgrounds,calc}

\title{\LARGE \bf
Temporal Tactile Encoding and Compliance for Intent-Aware Robot-to-Human Bimanual Handover
}

\author{
Pasquale Marra$^{1,2,\ast}$,
Stefano Berti$^{1,\ast}$,
Gabriele M. Caddeo$^{1}$
and Lorenzo Natale$^{1}$
}

\begin{document}

\twocolumn[{%
\renewcommand\twocolumn[1][]{#1}%
\maketitle
\begin{center}
    \vspace{-0.1in}
    \centering
    \captionsetup{type=figure}
    \includegraphics[width=1.0\linewidth]{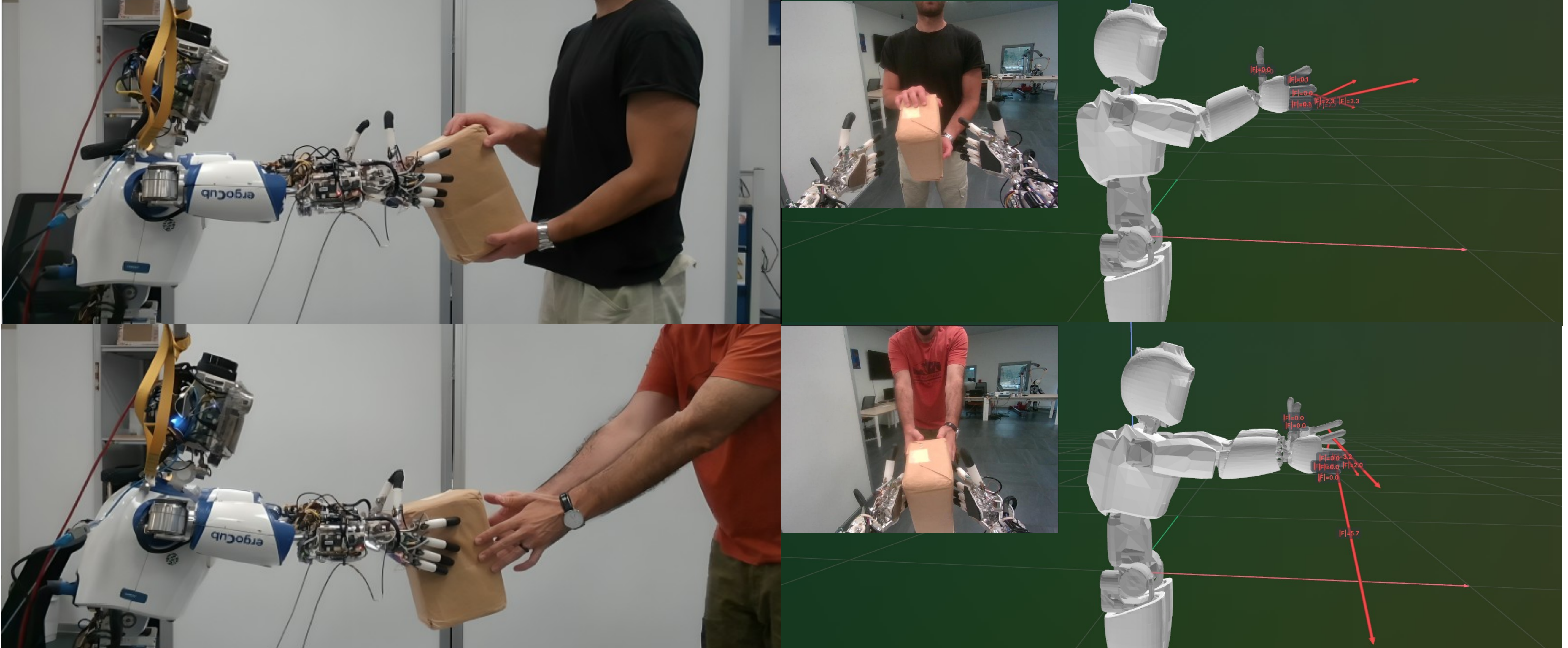}
    \captionof{figure}{Examples of release and hold behavior during handover. The left column shows the external view; the right shows the corresponding virtual view, with fingertip forces in red and the robot's egocentric view. Top row: a sustained pull toward the human triggers release; the virtual view shows an earlier frame capturing the force that caused it. Bottom row: a downward, wrong-direction force is applied, and the robot correctly maintains its grasp.}
    \label{fig:first_image}
\end{center}
}]
\thispagestyle{empty}
\pagestyle{empty}

\begin{abstract}
Reliable robot-to-human handover requires the robot to infer when the person is ready to receive the object and release it safely, comfortably, and at the right time. This is challenging because visual observations alone may not disambiguate clear taking intent from accidental contact, weak grasping, wrong-direction forces, or transient interactions. In this work we treat human-robot handover as an intrinsically multimodal problem. Our approach couples a VLA model with a compliance controller that reduces interaction forces during object transfer. We finetune the VLA model with human demonstrations using RGB observation, temporally encoded tactile feedback and proprioception.  
We evaluate the complete system in a human-subject study against two baselines: one without tactile feedback and one using tactile feedback without compliance control. We hypothesize that combining compliance and temporal tactile encoding yields the most reliable and comfortable handovers, as compliance facilitates physical interaction while tactile history captures sustained taking intent. Performance is measured through objective metrics and an ad-hoc questionnaire. The results show that the two components provide complementary benefits and substantially outperform the baselines. Project page: https://hsp-iit.github.io/temporal-tactile-bimanual-handover/.
\end{abstract}

\par
\begingroup
\footnotesize
\setlength{\parindent}{0pt}
\setlength{\parskip}{0pt}

\textsuperscript{1}Humanoid Sensing and Perception,
Istituto Italiano di Tecnologia, Genoa, Italy.
\textsuperscript{2}DIBRIS, Universit\`a di Genova, Genoa, Italy.
\textsuperscript{*}These authors contributed equally.
\par
This work was supported by the National Institute for Insurance
against Accidents at Work (INAIL), project ergoCub-core.
\par
Data collection complied with the ethical standards of the
2013 Declaration of Helsinki and was approved by
Comitato Etico Regione Liguria.
\par
\endgroup

\section{Introduction}

Robot-to-human handover is a fundamental capability for assistive, collaborative, and service robots \cite{10.5898/JHRI.2.1.Strabala,DUAN2024100145}. In a successful handover, the robot must coordinate physical transfer with the human receiver's intention and release the object at the appropriate moment \cite{10.5898/JHRI.2.1.Strabala,10.1145/2157689.2157692}. Releasing too early may lead to drops or unsafe interactions, while releasing too late forces the human to pull against the robot, reducing comfort and fluency \cite{10.1145/2157689.2157692}. Inferring taking intent is challenging because it depends on multiple cues, including visual motion, contact, grasp formation, pulling force, and load transfer. Vision alone may therefore be insufficient to distinguish a deliberate grasp-and-pull action from accidental contact, wrong-direction forces, or transient interactions \cite{li2026activecontactsensingrobust} (Fig.~\ref{fig:first_image}).

Tactile sensing has long been used in robotics~\cite{11162616}, both alone and with vision, to estimate contact-level signals such as force, shear, and slip~\cite{11127723, ford2025shearbasedgraspcontrolmultifingered, 10598389}, as well as object-level properties including shape, texture and pose~\cite{neuralfeels, 10160359, 9197046, 10.1007/978-3-032-37138-6_35}. These signals are especially important in dynamic, contact-rich tasks, where vision alone cannot directly observe occluded contacts or measure interaction forces. Robot-to-human handover is one such task: the robot must determine whether the human has established contact, is pulling in the intended direction, and is actively taking the object. Yet, how to integrate tactile feedback into learned handover policies remains an open question. Existing systems often rely on engineered release rules, while visuo-tactile learning has focused mainly on robot-object manipulation. Human-facing handover instead requires reliable, comfortable, and safe inference of taking intent~\cite{ortenzi2021handover}.

Compliance is important in handover because it allows the robot to yield during pulling, reducing resistance and improving physical fluency. It also limits interaction forces during unintended or prolonged contact, smoothing the exchange and reducing the risk of damage. However, compliance can introduce ambiguity for learned policies, since robot motion caused by yielding may become correlated with release. Tactile feedback helps resolve this ambiguity by distinguishing sustained taking intent from motion induced by the robot's own compliant response, while providing relevant information about human contact and grasp dynamics.

In this work, we approach robot-to-human package handover as an intrinsically multimodal problem that combines vision, touch, proprioception, and compliant behavior. Specifically, we investigate multimodal imitation learning~\cite{ARGALL2009469,Osa_2018} with a humanoid robot using NVIDIA GR00T 1.5 3B~\cite{gr00tn1_2025}, a VLA foundation model designed for post-training on specific embodiments and tasks. From demonstrations collected with 8 human subjects, we fine-tune the model using RGB observations, tactile feedback, and proprioception. Because the policy runs at 10 $Hz$ while the tactile sensors provide higher-frequency measurements, we encode approximately one second of tactile history to capture the evolution of contact and pulling behavior instead of relying only on instantaneous tactile readings. We evaluate three handover configurations: a full multimodal policy combining RGB observations, temporal tactile encoding, and compliance control; a no-tactile variant; and a no-compliance variant. This ablation tests whether tactile history and compliance provide complementary benefits for reliable and comfortable handover. We hypothesize that the full configuration performs best, since compliance facilitates the physical exchange while temporal tactile feedback provides evidence of sustained human taking intent. This design also reflects a hierarchical view of handover: GR00T N1 separates high-level vision-language processing from fast action generation, while in our setting compliance acts as an even faster low-level reflex layer beneath the learned policy. We test the learned policies with 10 human participants using both release-expected interactions and hold-expected conditions designed to probe premature or unsafe release. The evaluation combines objective handover metrics with subjective ratings of safety, smoothness, responsiveness, reliability, comfort, and overall preference, assessing both task completion and perceived interaction quality.

The contributions of this work are threefold:

\begin{itemize}
    \item We present a multimodal imitation-learning approach for humanoid robot-to-human package handover using RGB observations, fingertip tactile feedback, temporal tactile encoding, and compliance control.

    \item We study the roles of tactile history and compliance control through an ablation comparing a full multimodal policy against no-tactile and no-compliance variants.

    \item We evaluate the learned policies with human subjects, combining objective handover metrics with subjective ratings of safety, comfort, reliability, responsiveness, and preference.
\end{itemize}

\section{RELATED WORK}
Bimanual manipulation research has strong model-based foundations for force distribution, coordination, and safe physical interaction: constrained optimization for multifingered grasp-force distribution \cite{6289375}, decentralized admittance strategies for internal-wrench robustness \cite{11245924}, and impact-aware dual-arm grasping \cite{vansteen2023dualarmimpactawaregrasping}. These methods give strong stability and safety guarantees, but rely on explicit control laws rather than learning release behavior from multimodal demonstrations.

Visuo-tactile learning has shown strong potential for contact-rich bimanual manipulation: TactileAloha improves contact-sensitive assembly tasks such as zip-tie insertion and Velcro fastening \cite{11063285}; Bi-Touch and VT-Refine transfer tactile policies from simulation to real bimanual pushing, reorientation, and assembly \cite{lin2023bitouchbimanualtactilemanipulation,huang2025vtrefinelearningbimanualassembly}; BiTLA adds language-conditioned tactile-action modeling \cite{10.1145/3728485.3759237}. These works address robot–object manipulation rather than human-facing handovers, where the challenge lies not only in contact-rich control but also in determining when to release the object based on whether the human has clearly taken hold of it.

Human-robot handover research has studied release timing, intent prediction, and multimodal interaction \cite{ortenzi2021handover}, from timing and role adaptation in dynamic exchange \cite{Duan2024HumanrobotOH} to whole-body mobility, vision-based affordance, probabilistic motion prediction, and soft-hand interaction \cite{tulbure2026taskorientedrobothumanhandoverslegged,10355704,9812465,10777566}. Closest to our setting on the learning side, \cite{kim2025learningdynamichandover} learn dynamic robot-to-human handover motions from human-human demonstrations and pair them with impedance control for safe physical interaction, showing that learned, adaptive delivery improves handover time and user comfort over static baselines. These works give useful insight into handover fluency and delivery motion, but they decide \emph{where and how to move}, not \emph{when to let go}: the release decision still rests on visual, kinematic, or engineered features rather than the integrated feedback from vision and dense fingertip tactile histories learned directly by the policy.

A smaller body of work explicitly uses contact sensing for the release decision itself: tactile feedback for grasp stability and direction-gated release, tactile-glove-based object classification for coordinated multi-arm exchange \cite{9319192,MAZHITOV2023104311}, and, closest in intent to ours, proprioceptive filtering of a wrist force/torque signal combined with a bio-inspired adaptive release law that improves release fluency in human-subject handovers \cite{penzotti2025releasefluency}. These approaches typically rely on rule-based release logic or specialized controllers; we instead evaluate whether multimodal information, and especially dense fingertip tactile streams, including a temporal encoding, improve learned handover policies.

Recent humanoid foundation models offer a new way to learn such policies from demonstration. NVIDIA Isaac GR00T N1 is a generalist vision-language-action model for humanoid skills, trained on heterogeneous robot, human-video, and synthetic data and post-trainable for new embodiments and tasks through a dual-system architecture pairing slow vision-language reasoning with fast real-time action \cite{gr00tn1_2025}. Such models are a powerful backbone for humanoid manipulation, but the role of high-frequency fingertip tactile feedback in human-facing handover remains underexplored.

Our work targets this gap with multimodal imitation learning for humanoid robot-to-human handover, evaluating a full model (RGB, temporal tactile encoding, compliance control) against no-tactile and no-compliance variants. This isolates the contributions of tactile history and compliance while keeping the human-subject evaluation focused. Unlike prior tactile handover systems built on predefined release rules or specialized controllers, and unlike visuo-tactile bimanual work focused on robot-object tasks, we evaluate learned release behavior in real human-subject handovers.

\begin{figure*}
    \centering
    \vspace{0.2cm}
    \includegraphics[width=\textwidth]{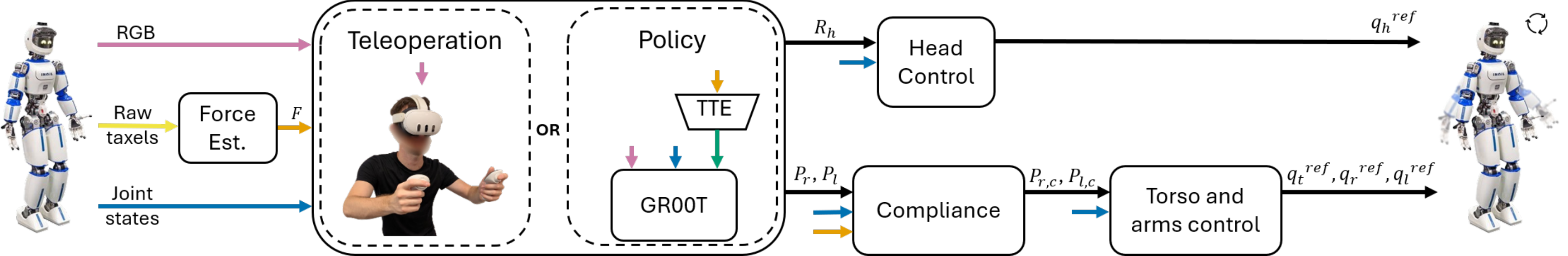}
    \caption{System overview. The robot's camera, fingertip forces $F$ (estimated from the raw Xela taxels), and joint states feed either the teleoperation interface (Meta Quest~3, during demonstration collection) or the learned policy (GR00T with the temporal tactile encoder TTE, at evaluation time). Both produce the same references: a desired head orientation $R_h$, tracked by the head controller ($q_h^{ref}$), and Cartesian hand poses $P_r$, $P_l$. The compliance module converts the hand references into compliant references $P_{r,c}$, $P_{l,c}$ based on the measured fingertip forces (Eq.~\ref{eq:compliance}); the torso-and-arms controller then computes the torso and arm joint references $q_t^{ref}$, $q_r^{ref}$, $q_l^{ref}$ sent to the robot.\label{fig:pipeline}}
    \vspace{-0.2cm}
\end{figure*}

\section{METHOD}

\subsection{Task and Platform}
We consider a bimanual robot-to-human package handover task on the humanoid robot ergoCub~\cite{sartore_towards_2026}. The robot grasps and holds a box with both hands and must release it once the human receiver has taken over the load, using an egocentric RGB camera and fingertip tactile sensors on both hands. Demonstrations are collected through teleoperation, and evaluation is carried out with human-subject trials. Such interactions may correspond either to a deliberate object transfer or to misleading contact patterns that resemble one. We next describe the teleoperation system and data collection procedure (\ref{subsec:data-collection}), the processing of the tactile data (\ref{subsec:tactile}), the compliance controller (\ref{subsec:compliance}), and the policy architecture (\ref{subsec:policy}). Figure~\ref{fig:pipeline} shows the overview of the approach.

\subsection{Teleoperation System and Data Collection}\label{subsec:data-collection}
 
Demonstrations were collected using a custom Meta Quest~3 teleoperation application, which tracks the operator's hand poses and headset orientation in a fixed world frame. These measurements were converted into Cartesian references for the robot hands and an orientation reference for the head. Teleoperation commands, robot states, and visual observations were recorded at 10$\mathrm{Hz}$ with LeRobot~\cite{cadenelerobot} and stored in its standard dataset format. Raw Xela fingertip~\cite{xelaroboticsuscuIIT} measurements were recorded separately at approximately $70\,\mathrm{Hz}$, with timestamps for subsequent alignment.
For each of the 8 participants, we collected 50 demonstrations: 25 without compliance, used to train the corresponding baseline, and 25 with compliance, yielding approximately 50k frames per condition, summing all the participants. For each condition, the participants performed five sets of five episodes, separated by pauses of approximately 1.5\,$\mathrm{min}$ to slightly rotate the robot, rest the Xela sensors, and restart the force-estimation module for recalibration. The first set contained standard handovers. In the second, the participant first moved their hands near the package without contact and then pulled it toward themselves. In the third, they grasped the package and initially applied forces in incorrect directions before performing a genuine pull. The fourth introduced temporally discontinuous contacts and arbitrary force directions, followed by a genuine pull. The final set randomly combined the previous patterns and ended with a pull toward the participant. Throughout each interaction, the teleoperator continuously adjusted the robot's upper-body posture by bending and rotating the torso and neck to follow the participant's motion. The participant verbally indicated the onset of the genuine pull, and the teleoperator released the package approximately one second later. This delay allowed the participant to load the package and ensured that the Xela sensors captured the resulting taxel variations before release.

\subsection{Force Estimation and Tactile Temporal Encoding}\label{subsec:tactile}
Both hands are equipped with a custom version of Xela fingertip tactile sensors. In this work we consider only the sensors on the four fingers excluding the thumb for each hand, given that the thumbs do not come in contact with the object during the task. Each sensor comprises seven magnetic taxels that can displace in three directions under applied force. The resulting raw taxel measurements are processed by finger-specific force estimators to estimate the 3-axis contact force at each fingertip. To obtain these estimators, we first collect a stimulation dataset for each finger. A spherical 3D-printed indenter is mounted on the end-effector of an Omega.3 robot, which is position-controlled to apply a wide range of contact forces. A regression model is then trained for each sensor using the corresponding dataset, following the procedure described in \cite{marra2026multifingeredforceawarecontrolhumanoid}. 
We then compare two ways of exposing the estimated forces to the policy. \emph{Instantaneous force} uses only the current-frame force reading, represented as a 24-dimensional vector corresponding to 4 fingers, 2 hands, and 3 force axes. \emph{Temporal Tactile Encoding (TTE)} instead summarizes a short causal history of contact using a frozen autoencoder trained separately from the policy. Specifically, a $(64,2,12)$ tactile window, corresponding to 64 timesteps, 2 hands, and 12 force values per hand from the 4 fingers' 3-axis forces, covering approximately \SI{0.9}{\second} of history. This window is flattened and passed through a fully connected encoder to obtain a 24-dimensional embedding (same size as the raw forces). The autoencoder is trained with a reconstruction loss $\mathcal{L}_{\mathrm{rec}}$ and two auxiliary terms: a force-direction loss $\mathcal{L}_{\mathrm{dir}}$ and a delta-direction loss $\mathcal{L}_{\Delta\mathrm{dir}}$.
The autoencoder is trained with the combined objective

\begin{equation}
\mathcal{L}_{\mathrm{TTE}}
=
\lambda_{\mathrm{rec}}\mathcal{L}_{\mathrm{rec}}
+
\lambda_{\mathrm{dir}}\mathcal{L}_{\mathrm{dir}}
+
\lambda_{\Delta}\mathcal{L}_{\Delta\mathrm{dir}} 
\label{eq:tte-objective}
\end{equation}

The two auxiliary losses are defined as:

\begin{equation}
\begin{aligned}
\mathcal{L}_{\mathrm{dir}}
&=
\frac{1}{|\mathcal{V}|}
\sum_{i \in \mathcal{V}}
d_{\mathrm{cos}}
\left(
\hat{\mathbf{f}}_{i},
\mathbf{f}_{i}
\right), \\
\mathcal{L}_{\Delta\mathrm{dir}}
&=
\frac{1}{|\mathcal{V}_{\Delta}|}
\sum_{i \in \mathcal{V}_{\Delta}}
d_{\mathrm{cos}}
\left(
\Delta\hat{\mathbf{f}}_{i},
\Delta\mathbf{f}_{i}
\right),
\end{aligned}
\label{eq:direction-losses}
\end{equation}

with
\begin{equation}
    \begin{aligned}
        \Delta\mathbf{f}_{t}&=\mathbf{f}_{t+d}-\mathbf{f}_{t},\\
\Delta\hat{\mathbf{f}}_{t} &= \hat{\mathbf{f}}_{t+d}-\hat{\mathbf{f}}_{t}
    \end{aligned}
\end{equation}
Here, $\mathbf{f}_{i}\in\mathbb{R}^{3}$ and
$\hat{\mathbf{f}}_{i}\in\mathbb{R}^{3}$ denote the target and
reconstructed force vectors, respectively, and $d_{\mathrm{cos}}(\mathbf{a},\mathbf{b})$ is the cosine distance between two vectors. Before computing the directional losses, both target and reconstructed forces are denormalized using the training-set statistics, so that the losses are evaluated in the original force space. The index $i=(b,t,h,k)$ is a compact multi-index, where $b$ denotes the sequence in the mini-batch, $t$ the time step, $h$ the hand, and $k$ the finger. In Eq.~\ref{eq:direction-losses},
the sets $\mathcal{V}$ and $\mathcal{V}_{\Delta}$ contain only indices
for which the magnitudes of the target force and target force difference,
respectively, exceed \SI{0.2}{\newton}. This excludes near-zero vectors,
for which direction is not well defined. Finally, $\lambda_{\mathrm{rec}}$ ($0.25$), $\lambda_{\mathrm{dir}}$ ($2$), $\lambda_{\Delta}$ ($1$) weight
the losses, and $d$ ($8$) is the temporal interval used to compute the force differences.

\subsection{Compliance and upper body control}\label{subsec:compliance}
Independently of the tactile representation, the robot's torso and arms controller (a second order inverse kinematics algorithm), by means of the compliance module, yields in the direction of external forces, letting the human's motion carry the robot's hands rather than resisting it. The compliance module receives the reference positions of the right and left hands, together with the fingertip forces, and computes compliant hand positions and velocities (Fig.~\ref{fig:pipeline}). Compliance is enforced according to:

\begin{equation}\label{eq:compliance}
 K_p \cdot (P_{h,c} - P_h) + K_d \cdot (\dot{P}_{h,c} - \dot{P}_h) = \sum_{i=0}^{4} F_{hi} \quad h = l, r\\
\end{equation}

where: $K_p, K_d > 0$ are scalar gains (in our case $K_p=50$, $K_d =40$), $P_h, \dot{P}_h\in\mathbb{R}^3$ $h=r,l$ denote the reference hand positions and velocities, $P_{h,c}, \dot{P}_{h,c}\in\mathbb{R}^3$ $h=r,l$ are the corresponding compliant positions and velocities, $F_{hi}$ is the force measured at the $i$-th fingertip of hand $h$.
The resulting compliant hand references are provided to the torso and arms controller that calculates the torso $q_t^{ref}$, right arm $q_r^{ref}$ and left arm $q_l^{ref}$ joint references for the robot. In parallel, the head controller regulates the desired head orientation \( R_h \), obtained either from teleoperation or from the policy, while compensating for torso motion, generating the head joint references $q_h^{ref}$.

\subsection{Policy}\label{subsec:policy}
All policies are built on NVIDIA Isaac GR00T-N1.5-3B \cite{gr00tn1_2025}, post-trained from its released checkpoint for our task with the prompt \emph{``Get the box and pass it to human''}. The policy receives egocentric RGB images together with a proprioceptive/tactile state vector and outputs a chunk of 16 Cartesian target actions for both hands and the head at  \SI{10}{\hertz}, corresponding to a \SI{1.6}{\second} prediction horizon. We execute the first 8 actions of each predicted chunk before querying the policy again. All ablated configurations share the same backbone, training recipe, and action space. More details are provided in Sec.~\ref{subsec:eval-human}.

\begin{figure}[t]
    \centering
    \vspace{0.2cm}
    \includegraphics[width=\columnwidth]{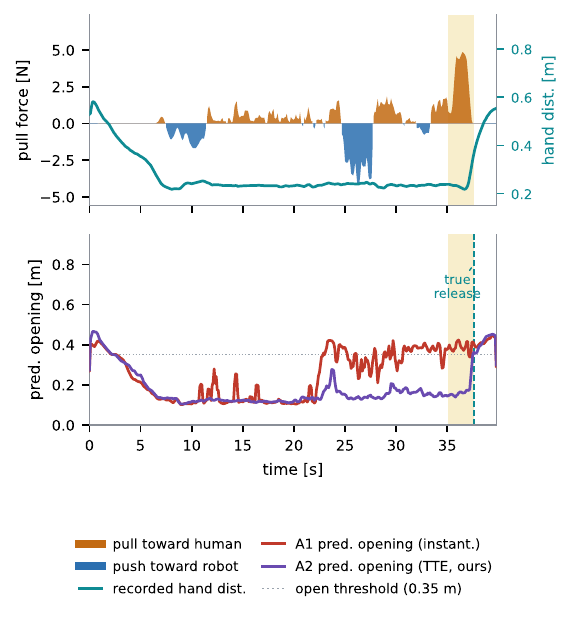}
    \caption{Challenging Test Set, episode 6 (repeated shakes and taps). Top: recorded pull force and hand distance. Bottom: the hand opening predicted by each policy in the final action of the 16-action chunk, with the \SI{0.35}{\meter} release threshold shown as a dotted line. The shaded region marks the final pull that triggers release. A1 incorrectly predicts release after non-causal pulls, whereas the TTE-based A2 detects it only during the final genuine pull.\label{fig:ablation}}
    \vspace{-0.2cm}
\end{figure}

\section{EXPERIMENTAL RESULTS}

\begin{figure*}
    \centering
    \vspace{0.2cm}
    \includegraphics[width=\textwidth]{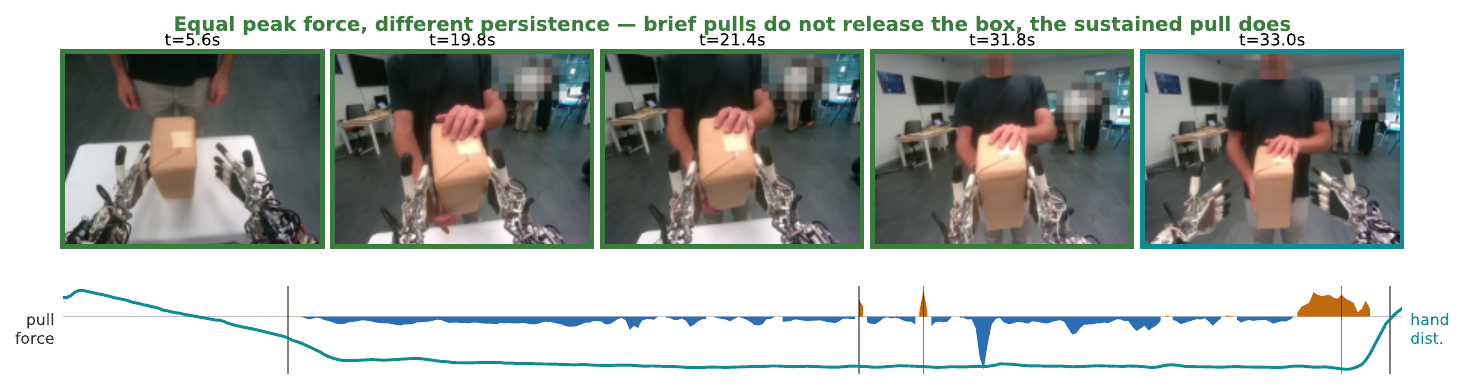}
    \caption{Full-system behavior (P1) in hold-expected trial T5. After grasping the box (first frame), the participant applies two brief pulls peaking at \SI{3.8}{\newton}, equal to the final genuine pull, but the robot keeps its hands closed (second and third frames). Sustaining the same force for $\sim$\SI{1.5}{\second}, rather than $\sim$\SI{0.2}{\second}, triggers release (fourth and final frames). The force and hand-distance plots show narrow transient peaks followed by a sustained pull, after which the hand distance increases. Thus, release depends on sustained pulling evidence rather than force magnitude alone. Faces are mosaicked for privacy.\label{fig:episode_shows}}
    \vspace{-0.2cm}
\end{figure*}

\subsection{Stage 1: Instantaneous Force vs.\ Temporal Tactile Encoding\label{subsec:ablation-results}}

As a first step, we validate the tactile representation used by the policy. This ablation keeps the human-subject evaluation tractable by avoiding an additional tactile-representation condition in the full study. We isolate the effect of tactile input by training two policies under identical conditions: \textbf{$A1$}, which receives instantaneous force measurements directly as observations, and \textbf{$A2$}, which receives the output of $TTE$. Both policies are trained from scratch with the same architecture, training data, and 100k-step budget.

We evaluate both policies open-loop on a held-out \emph{Challenging Test Set} containing 10 episodes and 3,776 frames. This set was recorded separately, it includes disturbances such as shakes, taps, and wrong-direction pushes, and was not used to train either policy. For each frame, we read out the predicted inter-hand distance as a proxy for the release/opening decision. Across the 10 episodes, $A1$ crosses the opening threshold 51 times, compared with only 3 times for $A2$. During the hold phase, $A1$ also produces 4 sustained false openings longer than $\SI{1.5}{\second}$, while $A2$ produces none. Since the two policies differ only in tactile representation, these results suggest that instantaneous force observations let transient contact fluctuations affect the release decision, whereas the causal history summarized by $TTE$ makes the policy require more sustained evidence before release. This supports using $TTE$ instead of raw instantaneous force in Stage 2.

Fig.~\ref{fig:ablation} shows the predicted hand opening for both policies on a representative Challenging Test Set episode, together with the recorded pull force and hand distance. Before the real release, marked by the dashed line, the human performs several correct-direction pulls that do not move the box, as indicated by the flat hand distance. $A1$ rises during these attempts and repeatedly commits to sustained false releases, showing that instantaneous force cannot reliably distinguish transient pulls from actual taking. In contrast, $A2$ remains nearly flat until the final pull, highlighted by the shaded span, when the release is genuinely under way.

\begin{figure}[t]
    \centering
    \includegraphics[width=\columnwidth]{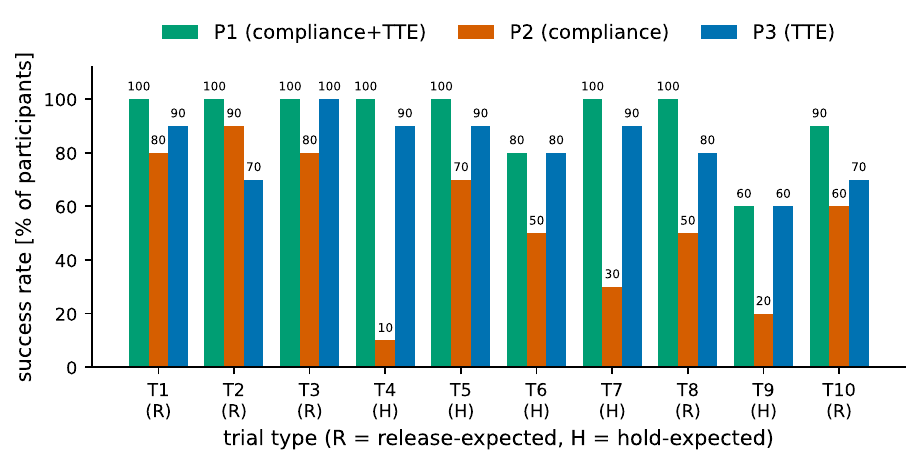}
    \caption{Success rate — participant got the expected outcome (release on R trials, hold on H trials) — for the three policies, broken down by all 10 trial types.\label{fig:bytrial}}
    \vspace{-0.2cm}
\end{figure}
\begin{table}[t]
\vspace{0.1cm}
\centering
\caption{Objective and subjective metrics, 10 participants $\times$ 10 trials per policy. \label{tab:metrics}}
\renewcommand{\arraystretch}{1.05}
\setlength{\tabcolsep}{3pt}
\footnotesize
\begin{tabular}{l r r r}
\toprule
\textbf{Metric} & \textbf{P1} & \textbf{P2} & \textbf{P3} \\
\cmidrule(lr){2-2}\cmidrule(lr){3-3}\cmidrule(lr){4-4}
 & compliance+TTE & compliance & TTE \\
\midrule
\multicolumn{4}{@{}l}{\textit{Objective -- 300 scored trials}} \\[1pt]
Success [\%] \ensuremath{\uparrow}                 & \bfseries 93 & 54 & 82 \\
Success, $T_r{=}2.8$s [\%] \ensuremath{\uparrow}     & \bfseries 92 & 45 & 80 \\
Release delay, median [s] \ensuremath{\downarrow}      & \bfseries 1.70 & 2.35 & 2.00 \\
Peak pull force [N] \ensuremath{\downarrow} & \bfseries 4.07 & 4.67 & 4.74 \\
\midrule
\multicolumn{4}{@{}l}{\textit{Subjective -- 10 post-block questionnaires}} \\[1pt]
Satisfaction, 1--7 Likert \ensuremath{\uparrow}    & \bfseries 6.2 & 3.3 & 4.7 \\
Preferred overall [/10] \ensuremath{\uparrow}     & \bfseries 9   & 1   & 0 \\
Rated safest [/10]  \ensuremath{\uparrow}         & \bfseries 10  & 0   & 0 \\
\bottomrule
\end{tabular}
\normalsize
\label{tab}

\end{table}
\subsection{Stage 2: Human-Subject Evaluation}\label{subsec:eval-human}
\subsubsection{Baselines}
Having selected TTE as the tactile representation (Sec.~\ref{subsec:ablation-results}), we train two additional policies to isolate the contributions of tactile history and compliance control. This results in three policies: \textbf{P1} (TTE + compliance), the full system; \textbf{P2} (compliance only), which removes TTE to evaluate the contribution of tactile history; and \textbf{P3} (TTE only), which removes compliance to evaluate the contribution of compliance control. Comparing P1 with P2 isolates the effect of tactile history under compliant interaction, while comparing P1 with P3 isolates the effect of compliance when temporal tactile information is available. Comparing P2 with P3 further indicates which component has a stronger effect on handover performance and user experience in this setting.

To train P3 fairly, we collected a separate no-compliance dataset, as previously described in Sec.~\ref{subsec:data-collection}. The number of episodes was matched to the compliant dataset used for P1 and P2. Since episode lengths were not identical, we subsampled the compliant dataset to match the number of frames used to train the no-compliance policy.
\subsubsection{Protocol}
We recruited 10 participants (5 female, 5 male), none of whom participated in the demonstration collection, and they provided informed consent for data collection and use. Each participant interacted with all three policies, with policy order counterbalanced across participants. For each policy, participants performed the same 10-trial protocol: 5 release-expected ($R$) trials and 5 hold-expected ($H$) trials. 

The $R$ trials are interactions in which the robot is expected to release the package ($T_1,T_2,T_3,T_8,T_{10}$ in Fig.~\ref{fig:bytrial}). They include a standard grasp-and-pull handover, a repeated grasp-and-pull after placing the package back on the table, a weak or slow pull, and pull trials performed after the robot bends while following the human moving backward, either with or without the human already touching the package during the bending motion. These trials evaluate handover success, consistency, sensitivity to weak intent, and whether the robot releases only when the human actually pulls the package.

The $H$ trials are interactions in which the robot is expected to keep holding the package ($T_4,T_5,T_6,T_7,T_9$ in Fig.~\ref{fig:bytrial}). They include grasping without pulling, brief touches or taps, continuous forces in the wrong direction, grasping without pulling after robot bends, and short transient pulls. These trials evaluate robustness to accidental contact, wrong-force interactions, visual/proprioceptive distractions, and transient intents. The bending-related trials specifically test whether the policy over-relies on robot motion induced by the compliance controller: even without tactile encoding, the policy may indirectly infer interaction through compliant bending and associate it with human pulling.

This protocol gives a total of 100 trials per policy: in $R$ trials, success requires the robot to release after a clear pull or load-transfer action. In $H$ trials, success requires the robot not to release during the initial \SI{30}{\second} disturbance phase. After this phase, participants are asked to clearly take the package so that the trial can be safely completed; therefore, every trial also contains a release attempt.

As objective metrics, we report both trial-level success (Success [\%]) and release-level timing performance (Success, $T_r{=}2.8$s [\%]). Trial-level success follows the expected outcome of each trial: release in $R$ trials and hold during the disturbance phase in $H$ trials. Timing-constrained success is instead computed over all release attempts, including both $R$ trials and the final release phase of $H$ trials. A release attempt is successful if the robot releases within a delay threshold $T_r$ after pull onset. We set $T_r$ from the training demonstrations rather than choosing it manually. Pooling pull-onset-to-release delays from all released episodes used to train the policies, the detectable releases have mean (\SI{1.33}{\second}), median (\SI{1.20}{\second}), and standard deviation (\SI{0.75}{\second}). We therefore set $T_r = \mu + 2\sigma \approx \SI{2.8}{\second}$, covering the large majority of training release timings while allowing margin for slower and more variable pulls from naive participants. We also report median release delay and peak pull force before release to quantify responsiveness and human effort.

\begin{figure*}[t]
    \centering
    \includegraphics[width=0.85\textwidth]{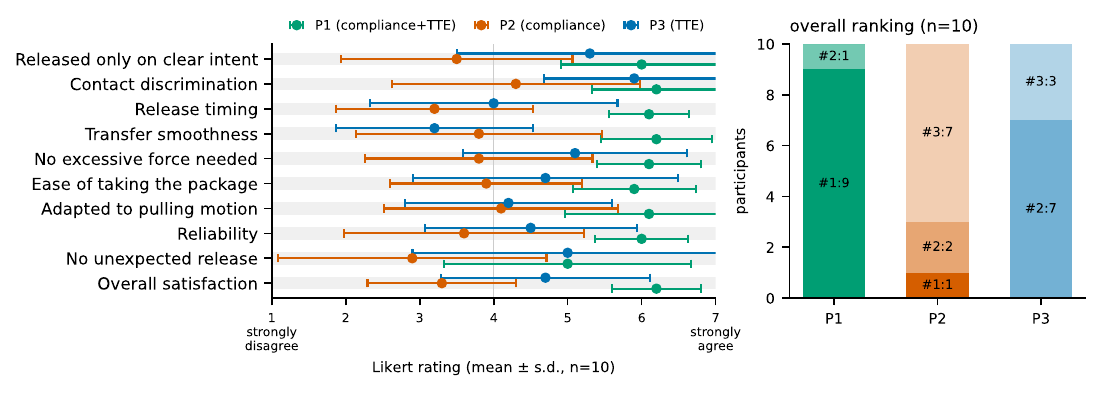}
    \caption{Left: Post-block questionnaire ratings (all 10 items, mean $\pm$ s.d.\ over 10 participants). The row titles summarize the meaning of the questions. Right: the overall best-to-worst ranking, represented as \#(rank position):(number of occurrences).\label{fig:likert}}
\end{figure*}

\subsubsection{Results}
Table~\ref{tab:metrics} summarizes the results across all 300 trials and 10 post-block questionnaires. P1 achieves the best overall performance. It reaches the highest raw success rate (93\%) and remains the most reliable configuration when release timing is considered, with 92\% success at $T_r=\SI{2.8}{\second}$, compared with 80\% for P3 and 45\% for P2. P1 also produces the fastest responses, with a median release delay of $\SI{1.70}{\second}$, compared with $\SI{2.00}{\second}$ for P3 and $\SI{2.35}{\second}$ for P2. Moreover, it requires the lowest peak pulling force from the human ($\SI{4.07}{\newton}$, versus $\SI{4.74}{\newton}$ for P3 and $\SI{4.67}{\newton}$ for P2). 

Figure~\ref{fig:bytrial} provides a breakdown by trial type. P1 achieves 100\% success on seven of the ten trial types, with lower performance only on T6 (80\%), T10 (90\%), and T9 (60\%). T9 is the most difficult condition for all configurations, as it contains impulsive pulls that can reach force magnitudes comparable to genuine taking actions, but remain shorter than the sustained pulls that should trigger release. This trial is therefore the human-subject counterpart of the transient-contact disturbances tested in the tactile-representation ablation. In one T9 trial for P1 and one T9 trial for P3, the participant applied an impulsive pull comparable to release-triggering interactions (approx. $\SI{2}{\second}$), leading to failure. Similarly, P1 fails once in T6 because the participant applied a continuous force in the human-taking direction, which triggered release despite the trial being hold-expected.

P2, which preserves compliance but receives no tactile input, performs worst on almost all objective metrics. Its limitations are especially evident on hold-expected trials, such as T4 (10\%, grasp without pulling) and T7 (30\%, transient pull). These trials are designed to test whether resting contact, brief pulling, or robot self-motion is mistaken for a genuine take. Without tactile information, motion caused by the robot's compliant response or by incidental human movements can be incorrectly interpreted as an intentional take. This confirms that compliance alone is not sufficient to reliably distinguish genuine taking intent from ambiguous or non-handover interactions. P3 retains TTE but removes compliance. Its success rate is lower than that of P1 (82\% versus 93\%), although it remains substantially more reliable than P2. This indicates that tactile history is the main factor enabling correct hold and release decisions. However, P3 requires higher pulling forces, responds more slowly, and performs worse on some release-expected trials, including T2 and T10. Without compliance, the robot does not yield to the human motion; consequently, participants sometimes pulled the package out of the robot's grasp before the hand-opening action was completed. This explains the occasional object drops observed only for P3 (4\% of its trials) and shows that compliance remains important for making the physical exchange smoother and more forgiving. Overall, TTE and compliance provide complementary benefits: TTE improves the release decision, while compliance improves the physical interaction. Figure~\ref{fig:episode_shows} illustrates this behavior for P1 during a T5 trial in which the brief pulls reach the same peak force ($\SI{3.8}{\newton}$) as the final genuine pull. The robot holds through the brief pulls and releases only on the sustained one, showing that the learned release decision does not reduce to a force-magnitude threshold. Instead, it depends on pulling evidence that is sustained over time and aligned with the taking direction. Examples of interactions for the different policies can be seen in the supplementary video.

Questionnaire results (Fig.~\ref{fig:likert}; 7-point Likert scale, 1 = strongly disagree, 7 = strongly agree) largely mirror the objective metrics. For the item ``the robot did not release unexpectedly,'' P1 and P3 obtain the same score (5.0), while P2 scores substantially lower (2.9), consistent with the lower false-release rates of the two policies using TTE. Similarly, for the ``robot distinguished accidental contact from deliberate taking``, P1 scores highest, with P3 close behind (6.2 vs.\ 5.9), while P2 scores lower (4.3), again suggesting that tactile history improves robustness to ambiguous contact. By contrast, smoothness and overall satisfaction separate P1 more clearly from the ablated variants. P1 obtains the highest smoothness score (6.2, compared with 3.8 for P2 and 3.2 for P3) and the highest overall satisfaction score (6.2, compared with 3.3 for P2 and 4.7 for P3). These results indicate that compliance contributes strongly to the subjective quality of the interaction, while tactile history alone is not sufficient to make the handover feel smooth and comfortable. The final ranking confirms the overall preference for the full system: 9 out of 10 participants ranked P1 first, and all 10 rated it as the safest policy. The only participant who ranked P1 second placed P2 first and explicitly favored P2's smoothness over P1's reliability in the free-text comments, suggesting that individual users may trade off comfort and reliability differently even when the aggregate preference strongly favors P1.

\section{CONCLUSION}

We studied two design choices for a learned humanoid handover policy: how tactile feedback is represented and whether compliance control is used during the physical exchange. In a two-stage evaluation, temporal tactile encoding proved more robust than instantaneous force readings in preventing false releases, while combining it with compliance control produced the most reliable, responsive, and preferred handover behavior. Tactile history accounts for most of the improvement in release robustness, whereas compliance further improves comfort and smoothness, showing that the two components are complementary rather than redundant.

Our data also suggest that compliance and tactile history operate on different timescales. The compliance controller reacts to human pulling within a low-level control cycle, immediately reducing resistance, while the learned release decision follows after TTE integrates tactile evidence over its temporal window. This suggests a hierarchy for human-facing manipulation: fast compliance control provides an immediate physical response, while the learned policy uses temporally integrated tactile feedback for reliable release decisions. Consistent with GR00T N1’s dual-system separation between slower reasoning and faster action generation \cite{gr00tn1_2025}, future humanoid handover systems may benefit from an explicit reflex-like compliance layer beneath the learned policy. Future work will study failure recovery during ambiguous interactions, adaptive release strategies for different users, and generalization to unseen objects, grasps, and handover configurations.

\bibliographystyle{IEEEtran}
\bibliography{bibliography}

\end{document}